\pdfoutput=1

\documentclass[11pt]{article}

\usepackage[]{emnlp2021}

\usepackage{times}
\usepackage{latexsym}

\usepackage[T1]{fontenc}

\usepackage[utf8]{inputenc}

\usepackage{microtype}

\usepackage{graphicx}
\usepackage{amsmath}
\usepackage{amssymb}
\usepackage{amsthm}
\usepackage{booktabs}
\usepackage{algorithm}
\usepackage{algorithmic}
\usepackage{dsfont}
\usepackage{multirow}
\usepackage{bm}
\usepackage{pifont}
\usepackage{bbding}

\title{Rethinking the Video Sampling and Reasoning Strategies for \\Temporal Sentence Grounding}

\author{Jiahao Zhu$^{1*}$, Daizong Liu$^{2\dag*}$, Pan Zhou$^{1\dag}$, Xing Di$^{3}$, Yu Cheng$^{4}$, Song Yang$^{5}$, \\
{\bf Wenzheng Xu$^{6}$, Zichuan Xu$^{7}$, Yao Wan$^{8}$, Lichao Sun$^{9}$, Zeyu Xiong$^{1}$} \\
$^{1}$Huazhong University of Science and Technology $^{3}$ProtagoLabs Inc $^{4}$Microsoft Research \\
$^{2}$Peking University $^{5}$Beijing Institute of Technology $^{6}$School of Sichuan University \\
$^{7}$Dalian University of Technology $^{9}$Lehigh University \\
$^{8}$School of Computer Sci. \& Tech., Huazhong University of Science and Technology \\
{\small
\texttt{\{jiahaozhu, panzhou, wanyao, zeyuxiong\}@hust.edu.cn, dzliu@stu.pku.edu.cn}} \\ 
{\small\texttt{xing.di@protagolabs.com, yu.cheng@microsoft.com, S.Yang@bit.edu.cn,}} \\
{\small\texttt{wenzheng.xu@scu.edu.cn, z.xu@dlut.edu.cn, lis221@lehigh.edu}}} 
\begin{document}
\maketitle
\footnote{$^{*}$Equal contributions. \quad $^{\dag}$Corresponding author.}
\vspace{40pt} \\
\begin{abstract}
Temporal sentence grounding (TSG) aims to identify the temporal boundary of a specific segment from an untrimmed video by a sentence query.
All existing works first utilize a sparse sampling strategy to extract a fixed number of video frames and then conduct multi-modal interactions with query sentence for reasoning.
However, we argue that these methods have overlooked two indispensable issues:
1) Boundary-bias: The annotated target segment generally refers to two specific frames as corresponding start and end timestamps. The video downsampling process may lose these two frames and take the adjacent irrelevant frames as new boundaries.
2) Reasoning-bias: Such incorrect new boundary frames also lead to the reasoning bias during frame-query interaction, reducing the generalization ability of model.
To alleviate above limitations, in this paper, we propose a novel Siamese Sampling and Reasoning Network (SSRN) for TSG, which introduces a siamese sampling mechanism to generate additional contextual frames to enrich and refine the new boundaries. Specifically, a reasoning strategy is developed to learn the inter-relationship among these frames and generate soft labels on boundaries for more accurate frame-query reasoning. Such mechanism is also able to supplement the absent consecutive visual semantics to the sampled sparse frames for fine-grained activity understanding.
Extensive experiments demonstrate the effectiveness of SSRN on three challenging datasets.
\end{abstract}

\section{Introduction}
\begin{figure}[t!]
\centering
\vspace{75pt}
\includegraphics[width=0.5\textwidth]{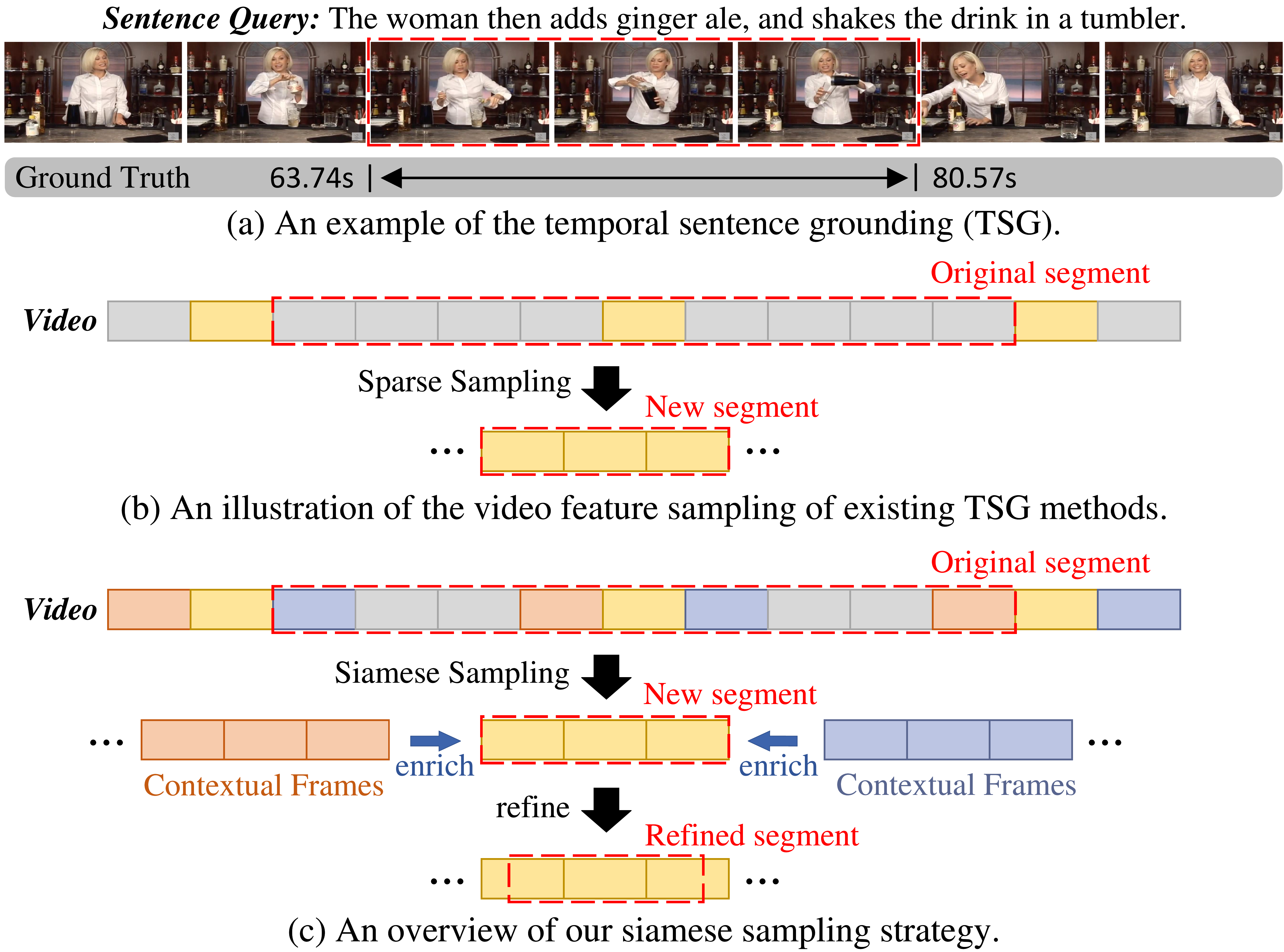}
\vspace{-20pt}
\caption{(a) An example of temporal sentence grounding task. (b) All existing TSG methods generally utilize a downsampling process to evenly extract a fixed number of frames from a long video. However, the new target segment is obtained by rounding operation and may introduces boundary bias since some original boundary frames are lost. (c) We propose a siamese sampling strategy to extract additional adjacent frames to enrich and refine the information of the sampled frames for generating more accurate boundary of the new segment.}
\label{fig:intro}
\vspace{-10pt}
\end{figure}

Temporal sentence grounding (TSG) is an important yet challenging task in natural language processing, which has drawn increasing attention over the last few years due to its vast potential applications in information retrieval \cite{2019Dual,yang2020tree} and human-computer interaction \cite{singha2018dynamic}. It aims to ground the most relevant video segment according to a given sentence query. As shown in Figure~\ref{fig:intro} (a), video and query information need to be deeply incorporated to distinguish the fine-grained details of adjacent frames for determining accurate boundary timestamps.

Previous TSG methods \cite{gao2017tall,chen2018temporally,zhang2019cross,yuan2019semantic,zhang2019learning,liu2018attentive,zhang2019man,liu2018cross,liu2021adaptive} generally follow an encoding-then-interaction framework that first extracts both video and query features and then conduct multi-modal interactions for reasoning. Since many videos are overlong while corresponding target segments are short, these methods simply utilize a sparse sampling strategy shown in Figure\ref{fig:intro} (b), which samples a fixed number of frames from each video to reconstruct a shorter video, and then learn frame-query relations for segment inferring. We argue that existing learning paradigm suffers from two obvious limitations:
1) Boundary-bias: Each video has a query-related segment, which refers to two specific frames as its start and end timestamps. Traditional sparse downsampling strategy extracts frames from videos with a fixed interval. A rounding operation is then applied to map the annotated segment to the sampled frames by keeping the same proportional length in both original and new videos. As a result, the ground-truth boundary frames may be filtered out and the query-irrelevant frames will be regarded as the actual boundaries, generating wrong labels for latter training.
2) Reasoning-bias: 
The query-irrelevant boundary frames in the newly reconstructed segment will also lead to incorrect frame-query interaction and reasoning in the training process, reducing the generalization ability of model.

To alleviate these two issues, a straightforward idea is to filter out the sampled boundary frames in the new segment if they are query-irrelevant. However, this will destroy the true segment length when we transfer the downsampled segment back to the original one during the inference process. Another straightforward idea is to directly keep the appropriate segment length (by float values) in the newly reconstructed video and then reason the query content in the new boundary to determine what percentage of this boundary is correct. However, the query-irrelevant boundaries lack sufficient query-related information for boundary reasoning. Based on the above considerations, we aim to extract additional frames adjacent to the sampled frames to enrich and refine their information for supplementing the consecutive visual semantics. In this way, the new boundary frames are well semantic-correlated to its original adjacent boundaries. Based on the refined boundary frames, we can keep and learn the appropriate segment length of the downsampled video for query reasoning.
Moreover, other inner frames are also enriched by their neighbors, capturing more consecutive visual appearances for fully understanding the entire activity.

Therefore, in this paper, we propose a novel Siamese Sampling and Reasoning Network (SSRN) for temporal sentence grounding task to generate additional contextual frames to enrich and refine the new boundaries. Specifically, we treat the sparse sampled video frames as anchor frames, and additionally extract several frames adjacent to each anchor frame as the siamese frames for semantic sharing and enriching. A siamese knowledge aggregation module is designed to explore internal relationships and aggregate contextual information among these frames. Then, a siamese reasoning module supplements the fine-grained contexts of siamese frames into the anchor frames for enriching their semantics. In this way, the query-related information are added into the new boundaries thus we can utilize an appropriate float value to represent the new segment length for query reasoning, addressing both boundary- and reasoning-bias. Moreover, other sampled frames are also equipped with more consecutive visual semantics from their original neighbors, which further benefits more fine-grained learning process. 

Our contributions are summarized as follows:
\begin{itemize}
    \item We propose a novel SSRN model which can sparsely extract multiple relevant frames from original videos to enrich the anchor frames for more accurate boundary prediction. To the best of our knowledge, we are the first to propose and address both boundary-bias and reasoning-bias in TSG task.
    \item We propose an effective siamese aggregation and reasoning method to correlate and integrate the contextual information of siamese frames to refine the anchor frames.
    \item Extensive experiments are conducted on three challenging public benchmarks, including ActivityNet Captions, TACoS and Charades-STA, demonstrating the effectiveness of our proposed SSRN method.
\end{itemize}

\section{Related Work}
Temporal sentence grounding (TSG) is a new task introduced recently \cite{gao2017tall,anne2017localizing}, which aims to localize the most relevant video segment from a video with sentence descriptions. All existing methods follow an encoding-then-interaction framework that first extracts video/query features and then conduct multi-modal interactions for segment inferring. 

Based on the interacted multi-modal features, traditional methods follow a \textit{propose-and-rank} paradigm to make predictions. Most of them \cite{ge2019mac, qu2020fine, xiao2021natural, liu2021adaptive,liu2021progressively,liu2020reasoning,liu2022learning,liu2022skimming,liu2022reducing,fang2022multi,liu2022few} typically utilize a proposal-based grounding head that first generates multiple candidate segments as proposals, and then ranks them according to their similarity with the query semantic to select the best matching one. Some of them \cite{gao2017tall,anne2017localizing} directly utilize multi-scale sliding windows to produce the proposals and subsequently integrate the query with segment representations via a matrix operation. 
To improve the quality of the proposals, latest works \cite{wang2019temporally,yuan2019semantic,zhang2019cross,cao2021pursuit,liu2021context,liu2020jointly,liu2022unsupervised,liu2022exploring,liu2022exploringa} integrate sentence information with each fine-grained video clip unit, and predict the scores of candidate segments by gradually merging the fusion feature sequence over time.

Recently, some \textit{proposal-free} works \cite{yuan2019find,wang2019language,rodriguez2020proposal,chenrethinking,mun2020local,zeng2020dense,zhang2020span,2021Parallel,nan2021interventional} directly predict the temporal locations of the target segment without generating complex proposals. These works directly select the starting and ending frames by leveraging cross-modal interactions between video and query. Specifically, they either regress the start/end timestamps based on the entire video representation \cite{yuan2019find,mun2020local}, or predict at each frame to determine whether this frame is a start or end boundary \cite{rodriguez2020proposal,chenrethinking,zeng2020dense,zhang2020span,2021Parallel}. 

Although the above two types of methods have achieved great performances, their video sampling strategy in encoding part is unreasonable that can lead to both boundary and reasoning bias. 
Specifically, the boundary bias is defined as the incorrect boundary of the new segment reconstructed by the video sparse sampling. The reasoning bias is defined as the incorrect correlation learning between the query-irrelevant frames and query.
In this paper, we aim to reduce the above bias by proposing a new siamese sampling and reasoning strategy to enrich the sampled frames and further refine the reconstructed segment boundary.

\begin{figure*}[t!]
\centering
\includegraphics[width=1.0\textwidth]{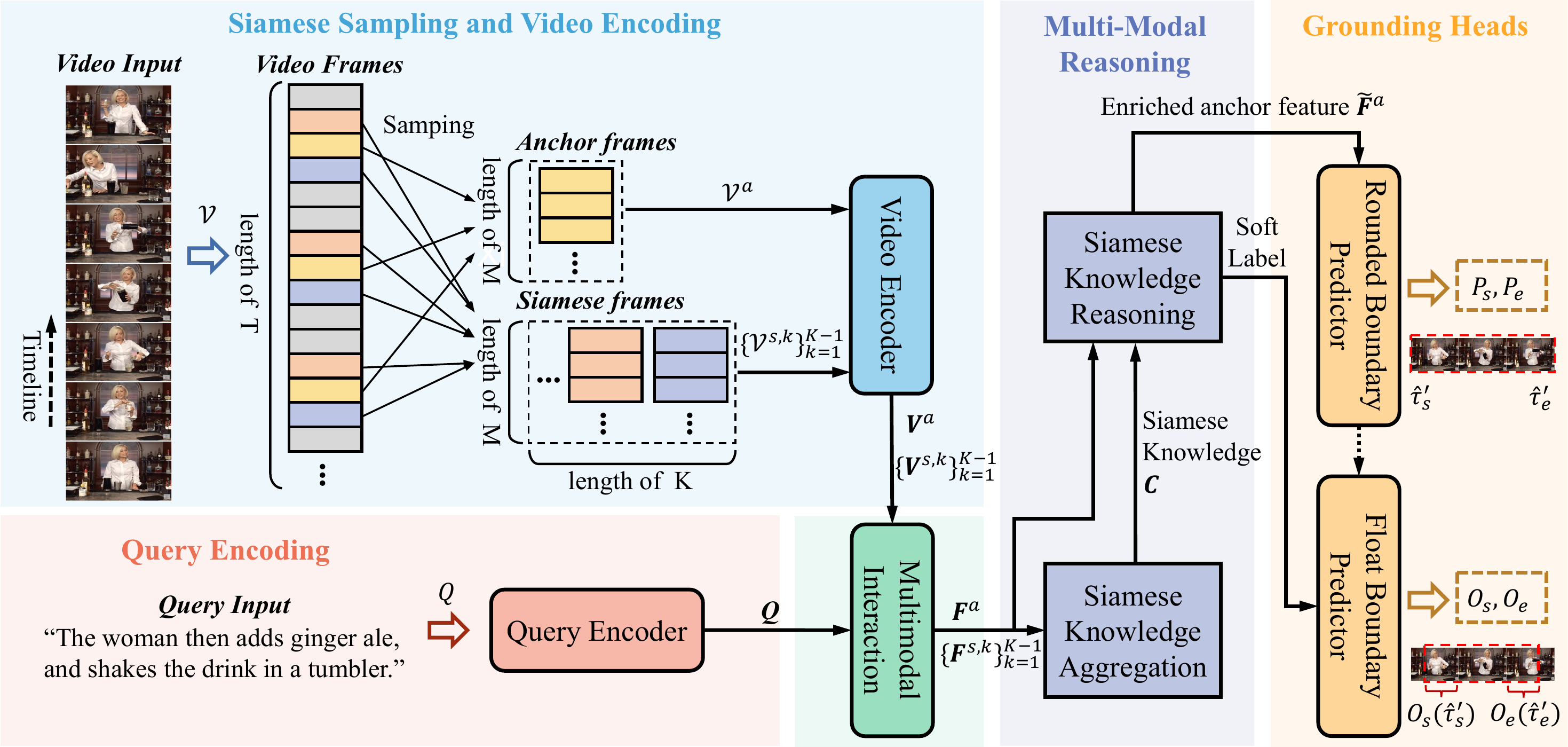}
\caption{Overview of our Siamese Sampling and Reasoning Network. Given a dense video, the anchor frames and siamese frames are first extracted by sparse sampling and siamese sampling, respectively. Then a video/query encoder and a multimodal interaction module are utilized to generate multimodal features. Next, a siamese knowledge generation module is proposed to model contextual relationship between anchor frames and siamese ones from the same video. After that, the siamese knowledge reasoning module exploits the siamese knowledge to enrich the information of the anchor frames for more accurate boundary prediction. At last, in the grounding heads, we utilize a soft label to learn more fine-grained boundaries of float value in addition to the rounded one.}
\label{fig:pipeline}
\vspace{-10pt}
\end{figure*}

\section{The Proposed Method}
Given an untrimmed video and a sentence query, we represent the video as $\mathcal{V}$ with a frame number of $T$. Similarly, the query with $N$ words is denoted as $\mathcal{Q}$. Temporal sentence grounding (TSG) aims to localize a segment $(\tau_s, \tau_e)$ starting at timestamp $\tau_s$ and ending at timestamp $\tau_e$ in video $\mathcal{V}$, which corresponds to the same semantic as query $\mathcal{Q}$.

The overall architecture of the proposed Siamese Sampling and Reasoning Network (SSRN) method is illustrated in Figure~\ref{fig:pipeline}. The SSRN framework contains four main components: (1) Siamese sampling and encoding: We sparsely downsample each long video into the anchor frames, and a new siamese sampling strategy additionally samples their adjacent frames as siamese frames. A video/query encoder then extracts visual/query features from all sampled video frames and query sentence respectively.
(2) Multi-modal interaction: After that, we interact the query features with the visual features for cross-modal interaction.
(3) Multi-modal reasoning: Next, to supplement the knowledge of siamese frames into the anchor frames, a siamese knowledge aggregation module is developed to determine how much the information of siamese frames are needed to inject into the anchor ones. Then, a reasoning module is utilized to enrich the anchor frames with the aggregated semantic knowledge. In this way, the contexts of both new boundaries and other sparse frames are enriched and can better represent the full and consecutive visual semantics.
(4) Grounding heads with soft labels: At last, we employ the grounding heads with soft label to predict more accurate boundaries via float value to keep the appropriate segment length. 
We illustrate the details of each component in the following subsections.

\subsection{Siamese Sampling and Encoding}
Given the dense video input $\mathcal{V}$, previous works generally downsample each video into a new video of fixed length to address the problem of overlong video. Considering the existing boundary-bias, we propose a siamese sampling strategy to additionally extract contextual adjacent frames nearby each sampled frame to enrich its query-related information for better determining the accurate new boundary. Here, we call the downsampled frames and their contextual frames as anchor frames and siamese frames, respectively.
Specifically, as shown in Figure~\ref{fig:intro} (c), following previous works, we directly construct the anchor video $\mathcal{V}^a$ by sparsely and evenly sampling $M$ frames from dense video frames of length $T$ ($T$ is usually much greater than $M$). The new siamese videos are then captured at different beginning indices in the original video but next to the frames of the anchor video. The same sample interval is utilized for all frames.
After siamese sampling, we can obtain multiple siamese videos with same length and similar global semantics as the anchor video. We denote the new siamese videos as $\{\mathcal{V}^{s,k}\}_{k=1}^{K}$ where $K$ means the siamese sample number. 

Since we utilize the sampling strategy to process the dense video frames, the start/end time of the target segment in original video sequence needs to be accurately mapped to the corresponding boundaries in the new video sequence of $M$ frames. Following almost all previous TSG methods \cite{zhang2019cross, zhang2020span, liu2021adaptive}, the new start/end index is generally calculated by $\hat{\tau}_{s(e)} = \lfloor \tau_{s(e)}/T \times M \rfloor$, where $\lfloor \cdot \rfloor$ denotes the rounding operator.
During the inference, the predicted segment boundary index can be easily converted to the corresponding time in the dense video via $\tau_{s(e)}  = \hat{\tau}_{s(e)}/M \times T$.
However, the rounding operation may produce boundary bias that the new boundary frames are not semantically correlated to the query semantic. Therefore, we further generate a soft label $\tilde{\tau}_{s(e)} = \langle \tau_{s(e)}/T \times M \rangle$ as an additional supervision to keep the appropriate segment length during training, where $\langle \cdot \rangle$ denotes the float result.

\noindent \textbf{Video encoder}~~
For video encoding, we first extract frame features by a pre-trained C3D network \cite{tran2015learning}, and then add a positional encoding \cite{vaswani2017attention} to provide positional knowledge. Such position encoding plays a crucial role in distinguishing semantics at diverse temporal locations. Considering the sequential characteristic in videos, a Bi-GRU \cite{chung2014empirical} is further applied to incorporate the contextual information along time series. We denote the extracted video features of both anchor video and siamese video as $\bm{V}^a, \{\bm{V}^{s,k}\}_{k=1}^K \in \mathbb{R}^{M \times D}$, respectively.

\noindent \textbf{Query encoder}~~
For query encoding, we first extract word embeddings by the Glove model \cite{pennington2014glove}. We also apply positional encoding and Bi-GRU to integrate the sequential information within the sentence. The final feature of the query is denoted as $\bm{Q} \in \mathbb{R}^{N \times D}$.

\subsection{Multi-Modal Interaction}
After obtaining the video features $\bm{V}^a, \{\bm{V}^{s,k}\}_{k=1}^K$ and query feature $\bm{Q}$, we utilize a co-attention mechanism \cite{lu-etal-2019-debug} to capture the cross-modal interactions between them. Specifically, for each video feature $\bm{V} \in \{\bm{V}^a\} \cup \{\bm{V}^{s,k}\}_{k=1}^K$, we first calculate the similarity between $\bm{V}$ and $\bm{Q}$ as:
\begin{equation}
    \bm{S} = \bm{V}(\bm{Q}\bm{W}_S)^{\top} \in \mathbb{R}^{M \times N},
\end{equation}
where $\bm{W}_S \in \mathbb{R}^{D \times D}$ projects the query features into the same latent space as the video. Then, we compute two attention weights as:
\begin{equation}
\begin{aligned}
    &\bm{A} = \bm{S}_r (\bm{Q}\bm{W}_S) \in \mathbb{R}^{M \times D}, \\
    &\bm{B} = \bm{S}_r \bm{S}_c^{\text{T}} \bm{V} \in \mathbb{R}^{M \times D},
\end{aligned}
\end{equation}
where $\bm{S}_r$ and $\bm{S}_c$ are the row- and column-wise softmax results of $\bm{S}$, respectively. We compose the final query-guided video representation by learning its sequential features as follows:
\begin{equation}
    \bm{F} = \text{Bi-GRU}([\bm{V};\bm{A};\bm{V}\odot \bm{A};\bm{V}\odot \bm{B}]) \in \mathbb{R}^{M \times D},
\end{equation}
where $\text{Bi-GRU}(\cdot)$ denotes the Bi-GRU layers, $[;]$ is the concatenate operation, and $\odot$ is the element-wise multiplication. The output $\bm{F} \in \{\bm{F}^a\} \cup \{\bm{F}^{s,k}\}_{k=1}^K$ encodes visual features with query-guided attention.

\subsection{Multi-Modal Reasoning Strategy}
Note that the query-irreverent new boundary frames encoded in the anchor video feature $\bm{F}^a$ has insufficient query-guided visual information for latter boundary prediction.
To address this issue, we propose a new multi-modal reasoning strategy to enrich the query-related knowledge in anchor features $\bm{F}^a$ referring to the contextual information in siamese features  $\{\bm{F}^{s,k}\}_{k=1}^K$. In detail, the multi-modal reasoning strategy consists of two components: a siamese knowledge aggregation module and a siamese knowledge reasoning module.

\noindent \textbf{Siamese knowledge aggregation}~~
Intuitively, features with close visual-query correlation are expected to generate more consistent predictions of segment probabilities. To this end, we utilize a siamese knowledge aggregation module to generate interdependent knowledge from siamese features to anchor ones to enrich the contexts of anchor features and refine the prediction.

We propose to propagate and integrate knowledge between the query-guided visual features $\bm{F}^a$ and $\{\bm{F}^{s,k}\}_{k=1}^K$. Specifically, we first obtain their semantic similarities by calculating their pairwise cosine similarity scores as:
\begin{equation}
\bm{C}(i, k) = \frac{(\bm{F}_i^a)(\bm{F}_i^{s,k})^{\top}}{\parallel \bm{F}_i^a \parallel_2 \parallel \bm{F}_i^{s,k} \parallel_2},
\end{equation}
where $\bm{C} \in \mathbb{R}^{M \times K}$ is interdependent similarity matrix, $\parallel \cdot \parallel_2$ is $l_2$-norm, $i \in \{1,2,...,M\}$ is the indices of features and $k \in \{1,2,...,K\}$ is the indices of siamese videos. Here, each anchor frame is needed to be enriched by only its siamese frames. We employ a softmax function to each row of the similarity matrix $\bm{C}$ as:
\begin{equation}
\bm{C}(i, k) = 
\frac{exp(\bm{C}(i, k))}{\sum exp(\bm{C}(i, k))},
\end{equation}
where the new $\bm{C}$ indicates the contextual affinities between each anchor feature and its corresponding siamese features. 

\noindent \textbf{Siamese knowledge reasoning}~~
After that, we propose to adaptively propagate and merge the siamese knowledge into the anchor features for enriching the query-aware information. This is especially helpful when we determine more accurate boundaries for the downsampled video. Specifically, 
The integration process can be formulated as:
\begin{equation}
\widetilde{\bm{F}}^a = \sum_{k=1}^K \bm{C}(:,k) \cdot (\bm{F}^{s,k}\bm{W}_1) \in \mathbb{R}^{M \times D},
\end{equation}
where $\widetilde{\bm{F}}^a$ is the propagated semantic vector in anchor video. In order to avoid over propagation and involves in irrelevant noisy information, we further exploit a residual design with a learnable weight to enrich the anchor video as:
\begin{equation}
\widetilde{\bm{F}}^a = \alpha \sum_{k=1}^K \bm{C}(:,k) \cdot (\bm{F}^{s,k}\bm{W}_1) + (1 - \alpha)\bm{F}^a\bm{W}_2,
\end{equation}
where $\bm{W}_1, \bm{W}_2 \in \mathbb{R}^{D \times D}$ are projection matrices, weighting factor $\alpha \in [0, 1]$ is a hyper-parameter.
With the above formulations, the knowledge of the siamese samples within the same video can be propagated and integrated to the anchor one.

\subsection{Grounding Heads with Soft Label}
For the final segment boundary prediction, we first follow the span predictor in \cite{zhang2020span} to utilize two stacked-LSTM with two corresponding feed-forward layers to predict the start/end scores of each frame. In details, we send the contextual multi-modal feature $\widetilde{\bm{F}}^a \in \mathbb{R}^{M \times D}$ into this span predictor and apply the softmax function on its two outputs to produce the probability distributions $P_s,P_e \in \mathbb{R}^{M}$ of start and end boundaries. We utilize the rounded boundary $\hat{\tau}_{s(e)}$ to generate the coarse label vectors $Y_{s(e)}$ to supervise $P_s,P_e$ as:
\begin{equation}
\mathcal{L}_1 = f_{CE}(P_{s}, Y_{s}) + f_{CE}(P_{e}, Y_{e}),
\end{equation}
where $f_{CE}$ represents cross-entropy loss function.
The predicted timestamps $(\hat{\tau_{s}}', \hat{\tau_{e}}')$ are obtained from the maximum scores of start and end predictions $P_{s(e)}$ of frames as:
\begin{equation}
(\hat{\tau_{s}}', \hat{\tau_{e}}') = arg \max_{\hat{\tau_{s}}', \hat{\tau_{e}}'} P_s(\hat{\tau_{s}}') P_e(\hat{\tau_{e}}'),
\end{equation}
where $0 \le \hat{\tau}_{s}' \le \hat{\tau}_{e}' \le M$.

Since the above predictions are coarse on the segment boundaries with boundary-bias, we further utilize a parallel prediction head on $\widetilde{\bm{F}}^a$ to predict more fine-grained float boundaries on the downsampled boundary frames. Specifically, we utilize the float boundary $\tilde{\tau}_{s(e)}$ to generate the soft labels $Y_{s(e)}'$, and $\widetilde{\bm{F}}^a$ is fed into a single feed-forward layer to predict the float boundaries $O_{s(e)}$ supervised by our designed soft labels $Y_{s(e)}'$ as follows:
\begin{equation}
\mathcal{L}_2 = \mathcal{R}_1(O_{s(e)}-Y_{s(e)}'),
\end{equation}
where $\mathcal{R}_1$ is the smooth L1 loss. The final predicted segment is calculated by:
\begin{equation}
(\tilde{\tau}_s', \tilde{\tau}_e') = (\hat{\tau}_{s}'+1-O_s(\hat{\tau}_{s}'),\hat{\tau}_{e}'-1+O_s(\hat{\tau}_{e}')).
\end{equation}

\section{Experiments}
\subsection{Datasets and Evaluation}
\noindent \textbf{ActivityNet Captions}~~
This dataset \cite{krishna2017dense} contains 20000 untrimmed videos from YouTube with 100000 textual descriptions. The videos are 2 minutes on average, and the annotated video clips have significant variation of length, ranging from several seconds to over 3 minutes. Following public split, we use 37417, 17505, and 17031 sentence-video pairs for training, validation, and testing.

\noindent \textbf{TACoS}~~
TACoS \cite{regneri2013grounding} contains 127 videos. The videos from TACoS are collected from cooking scenarios, thus lacking the diversity. They are around 7 minutes on average. We use the same split as \cite{gao2017tall}, which includes 10146, 4589, 4083 query-segment pairs for training, validation and testing.

\begin{table}[t!]
    \small
    \centering
    \setlength{\tabcolsep}{1.2mm}{
    \begin{tabular}{c|c|cccc}
    \toprule
    \multirow{2}*{Method} & \multirow{2}*{Feature} & R@1, & R@1, & R@5, & R@5 \\ 
    ~ & ~ & IoU=0.5 & IoU=0.7 & IoU=0.5 & IoU=0.7  \\ \midrule 
    TGN & C3D & 28.47 & - & 43.33 & - \\
    CTRL & C3D & 29.01 & 10.34 & 59.17 & 37.54  \\
    QSPN & C3D & 33.26 & 13.43 & 62.39 & 40.78  \\
    CBP & C3D & 35.76 & 17.80 & 65.89 & 46.20  \\
    GDP & C3D & 39.27 & - & - & -  \\
    VSLNet & C3D & 43.22 & 26.16 & - & -  \\
    CMIN & C3D & 43.40 & 23.88 & 67.95 & 50.73 \\
    DRN & C3D & 45.45 & 24.36 & 77.97 & 50.30 \\ 
    2DTAN & C3D & 44.51 & 26.54 & 77.13 & 61.96 \\ 
    APGN & C3D & 48.92 & 28.64 & 78.87 & 63.19 \\
    MGSL & C3D & 51.87 & 31.42 & 82.60 & 66.71 \\
    \midrule
    \textbf{SSRN} & C3D & \textbf{54.49} & \textbf{33.15} & \textbf{84.72} & \textbf{68.48} \\ \bottomrule
    \end{tabular}}
    \caption{Performance compared with the state-of-the-art TSG models on ActivityNet Captions dataset.}
    \vspace{-10pt}
    \label{tab:compare}
\end{table}

\begin{table}[t!]
    \small
    \centering
    \setlength{\tabcolsep}{1.2mm}{
    \begin{tabular}{c|c|cccc}
    \toprule
    \multirow{2}*{Method} & \multirow{2}*{Feature} & R@1, & R@1, & R@5, & R@5,  \\ 
    ~ & ~ & IoU=0.3 & IoU=0.5 & IoU=0.3 & IoU=0.5 \\ \midrule 
    TGN & C3D & 21.77 & 18.90 & 39.06 & 31.02 \\
    CTRL & C3D & 18.32 & 13.30 & 36.69 & 25.42 \\
    QSPN & C3D & 20.15 & 15.23 & 36.72 & 25.30  \\
    CBP & C3D & 27.31 & 24.79 & 43.64 & 37.40  \\
    GDP & C3D & 24.14 & - & - & - \\
    VSLNet & C3D & 29.61 & 24.27 & - & - \\
    CMIN & C3D & 24.64 & 18.05 & 38.46 & 27.02 \\
    DRN & C3D & - & 23.17 & - & 33.36 \\ 
    2DTAN & C3D & 37.29 & 25.32 & 57.81 & 45.04  \\
    APGN & C3D & 40.47 & 27.86 & 59.98 & 47.12 \\
    MGSL & C3D & 42.54 & 32.27 & 63.39 & 50.13 \\
    \midrule
    \textbf{SSRN} & C3D & \textbf{45.10} & \textbf{34.33} & \textbf{65.26} & \textbf{51.85} \\ \bottomrule
    \end{tabular}}
    \caption{Performance compared with the state-of-the-art TSG models on TACoS datasets.}
    \label{tab:compare2}
\end{table}

\begin{table}[t!]
    \small
    \centering
    \setlength{\tabcolsep}{1.2mm}{
    \begin{tabular}{c|c|cccc}
    \toprule 
    \multirow{2}*{Method} & \multirow{2}*{Feature} & R@1, & R@1, & R@5, & R@5, \\ 
    ~ & ~ & IoU=0.5 & IoU=0.7 & IoU=0.5 & IoU=0.7 \\ \midrule 
    2DTAN & VGG & 39.81 & 23.25 & 79.33 & 51.15 \\
    APGN & VGG & 44.23 & 25.64 & 89.51 & 57.87 \\
    \textbf{SSRN} & VGG & \textbf{46.72} & \textbf{27.98} & \textbf{91.37} & \textbf{59.64} \\ \midrule
    CTRL & C3D & 23.63 & 8.89 & 58.92 & 29.57 \\
    QSPN & C3D & 35.60 & 15.80 & 79.40 & 45.40 \\
    CBP & C3D & 36.80 & 18.87 & 70.94 & 50.19 \\
    GDP & C3D & 39.47 & 18.49 & - & - \\
    APGN & C3D & 48.20 & 29.37 & 89.05 & 58.49 \\
    \textbf{SSRN} & C3D & \textbf{50.39} & \textbf{31.42} & \textbf{90.68} & \textbf{59.94} \\ \midrule
    DRN & I3D & 53.09 & 31.75 & 89.06 & 60.05 \\ 
    APGN & I3D & 62.58 & 38.86 & 91.24 & 62.11 \\
    MGSL & I3D & 63.98 & 41.03 & 93.21 & 63.85 \\
    \textbf{SSRN} & I3D & \textbf{65.59} & \textbf{42.65} & \textbf{94.76} & \textbf{65.48} \\ \bottomrule
    \end{tabular}}
    \caption{Performance compared with the state-of-the-art TSG models on Charades-STA datasets.}
    \label{tab:compare3}
    \vspace{-10pt}
\end{table}

\noindent \textbf{Charades-STA}~~
Charades-STA is built on the Charades dataset \cite{sigurdsson2016hollywood}, which focuses on indoor activities. 
The video length of Charades-STA dataset is 30 seconds on average, and there are 12408 and 3720 moment-query pairs in the training and testing sets, respectively.

\noindent \textbf{Evaluation}~~
Following previous works \cite{gao2017tall,liu2021adaptive}, we adopt “R@n, IoU=m” as our evaluation metrics. The “R@n, IoU=m” is defined as the percentage of at least one of top-n selected moments having IoU larger than m,  which is the higher the better.

\subsection{Implementation Details}
For video encoding, we apply C3D \cite{tran2015learning} to encode the videos on all three datasets, and also extract the I3D \cite{carreira2017quo} and VGG \cite{simonyan2014very} features on Charades-STA dataset for fairly comparing with other methods.
Following previous works, we set the length $M$ of the sampled anchor video sequences to 200 for ActivityNet Captions and TACoS datasets, 64 for Charades-STA dataset, respectively. As for sentence encoding, we utilize Glove word2vec \cite{pennington2014glove} to embed each word to a 300-dimension feature. The hidden state dimensions of Bi-GRU and Bi-LSTM are set to 512. 
The number $K$ of the sampled siamese frames for each anchor frame is set to 4.
We train our model with an Adam optimizer with leaning rate $8 \times 10^{-4}$, $3 \times 10^{-4}$, $4 \times 10^{-4}$ for ActivityNet Captions, TACoS, and Charades-STA datasets, respectively. 
The batch size is set to 64.

\subsection{Comparison with State-of-the-Arts}
\noindent \textbf{Compared methods}~~
We compare our SSRN with state-of-the-art methods, including: (1) \textit{propose-and-rank} methods: TGN \cite{chen2018temporally}, CTRL \cite{gao2017tall}, QSPN \cite{xu2019multilevel}, CBP \cite{wang2019temporally}, CMIN \cite{zhang2019cross}, 2DTAN \cite{zhang2019learning}, APGN \cite{liu2021adaptive}, MGSL \cite{liu2022memory}. 
(2) \textit{proposal-free} methods: GDP \cite{chenrethinking}, VSLNet \cite{zhang2020span}, DRN \cite{zeng2020dense}. 

\begin{table}[t!]
    \small
    \centering
    \setlength{\tabcolsep}{0.7mm}{
    \begin{tabular}{c|ccccccc}
    \toprule 
    ~ & CTRL & TGN & 2DTAN & CMIN & DRN & APGN & \textbf{SSRN }\\ \midrule
    VPS $\uparrow$ & 0.45 & 1.09 & 1.75 & 81.29 & 133.38 & 146.67 & \textbf{158.12} \\ \midrule
    Para. $\downarrow$ & \textbf{22} & 166 & 363 & 78 & 214 & 91 & 184 \\ \bottomrule
    \end{tabular}}
    \caption{Efficiency comparison in terms of video per second (VPS) and parameters (Para.).}
    \label{tab:efficient}
    \vspace{-10pt}
\end{table}

\noindent \textbf{Quantitative comparison}~~
As shown in Table \ref{tab:compare}, \ref{tab:compare2} and \ref{tab:compare3}, our SSRN outperforms all the existing methods by a large margin. Specifically, on ActivityNet Captions dataset, compared to the previous best method MGSL, we outperform it by 2.62\%, 1.73\%, 2.12\%, 1.77\% in all metrics, respectively. 
Although TACoS dataset suffers from similar kitchen background and cooking objects among the videos, it is worth noting that our SSRN still achieves significant improvements.
Compared to the previous best method MGSL, our method brings significant improvement of 2.06\% and 1.72\% in the strict ``R@1, IoU=0.5” and ``R@5, IoU=0.5” metrics, respectively.
On Charades-STA dataset, for fair comparisons with other methods, we perform experiments with same features (i.e., VGG, C3D, and I3D) reported in their papers. It shows that our SSRN reaches the highest results over all metrics. 

\noindent \textbf{Efficiency comparison}~~
To compare the efficiency of our SSRN with previous methods, we make a fair comparison on a single Nvidia TITAN XP GPU on the TACoS dataset. As shown in Table \ref{tab:efficient}, it can be observed that we achieve much faster processing speeds with a competitive model sizes. 

\begin{table}[t!]
    \small
    \centering
    \setlength{\tabcolsep}{0.8mm}{
    \begin{tabular}{c|ccccc|cc}
    \toprule 
    \multirow{2}*{Model} & \multirow{2}*{Anchor} & \multirow{2}*{Siamese} & \multirow{2}*{SKA} & \multirow{2}*{SKR} & \multirow{2}*{SL} & R@1, & R@1, \\ 
    ~ & ~ & ~ & ~ & ~ & ~ & IoU=0.5 & IoU=0.7 \\ \midrule 
    \ding{172} & $\checkmark$ & $\times$ & $\times$ & $\times$ & $\times$ & 42.78 & 26.35 \\ 
    \ding{173} & $\checkmark$ & $\times$ & $\times$ & $\times$ & $\checkmark$ & 43.64 & 26.81 \\ 
    \midrule
    \ding{174} & $\checkmark$ & $\checkmark$ & $\times$ & $\times$ & $\times$ & 45.50 & 27.93 \\
    \ding{175} & $\checkmark$ & $\checkmark$ & $\times$ & $\checkmark$ & $\times$ & 48.97 & 29.36 \\
    \ding{176} & $\checkmark$ & $\checkmark$ & $\checkmark$ & $\checkmark$ & $\times$ & 51.26 & 31.02 \\ 
    \midrule
    \ding{177} & $\checkmark$ & $\checkmark$ & $\checkmark$ & $\checkmark$ & $\checkmark$ & \textbf{54.49} & \textbf{33.15} \\
    \bottomrule
    \end{tabular}}
    \caption{Main ablation studies on ActivityNet Captions dataset, where ``Anchor" and ``Siamese" denote the anchor and siamese frames, ``SKA" and ``SKR" denote the siamese knowledge aggregation and siamese knowledge reasoning, ``SL" denotes the usage of soft label.}
    \label{tab:ablation1}
    \vspace{-10pt}
\end{table}

\begin{figure*}[t!]
\centering
\includegraphics[width=1.0\textwidth]{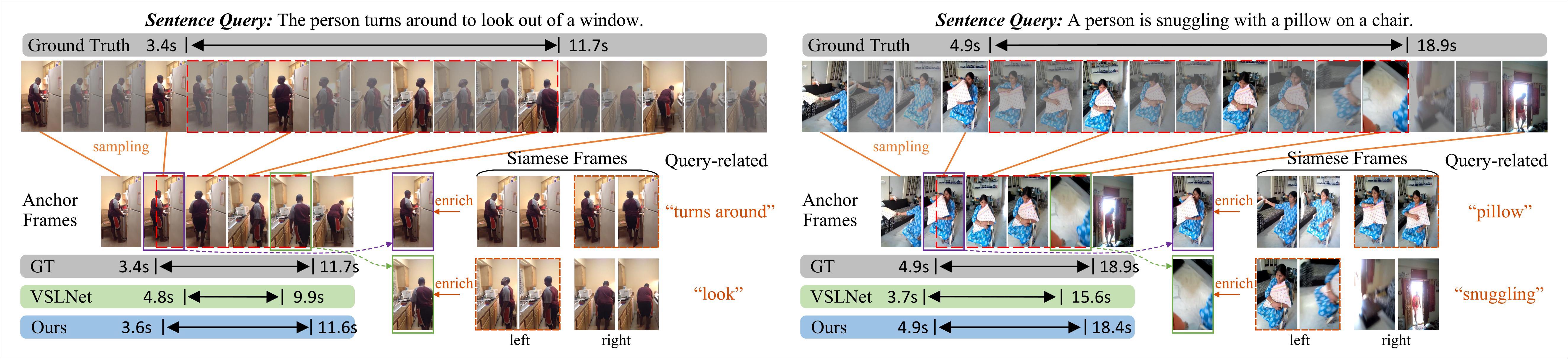}
\vspace{-20pt}
\caption{The visualization examples to show the benefits from the siamese frames. Due to the boundary-bias during the sparse sampling process, previous VSLNet method filters out the true-positive boundary frames and fails to predict the accurate boundaries. Instead, our siamese learning strategy supplements the query-related information of the adjacent frames into the ambiguous downsampled boundary-frames for predicting more precise boundaries.}
\label{fig:result}
\vspace{-10pt}
\end{figure*}

\subsection{Ablation Study}
\noindent \textbf{Effect of the siamese learning strategy}~~
As shown in Table~\ref{tab:ablation1}, we set the network without both siamese sampling/reasoning and soft label training as the baseline (model \ding{172}). Compared with the baseline, the model \ding{174} additionally extracts siamese frames for contextual learning, and can apparently improve the accuracy. It directly utilizes average operation to aggregate siamese knowledge and exploit concatenation for knowledge reasoning, which validates that multiple frames from same videos can really bring some strong knowledge to enhance the network. When further applying the SKR module on model \ding{174}, the model \ding{175} performs better, demonstrating the effectiveness of our SKR module.
When we further add the SKG module, our model \ding{176} can reach a higher performance, which can demonstrate the effectiveness of building the interdependent knowledge (i.e., siamese knowledge) for integrating the samples. It can also prove that adaptively reasoning by our siamese knowledge is better than the purely
average operation. We think that the siamese knowledge not only serves as the knowledge-routed representation, but also implicitly constrains the semantic consistency of frames in the space of frame-text features.

\noindent \textbf{Effect of the usage of soft label}~~
We also investigate whether our soft label (float value) of the segment boundary contributes to the performance of our model. As shown in Table~\ref{tab:ablation1}, directly applying the soft label learning to the baseline does not bring significant performance improvement (model \ding{173}). This is mainly because that the boundary frame may be query-irrelevant and its feature is not able to be accurately matched with the query. Instead, comparing model \ding{177} with model \ding{176}, model \ding{177} enriches the boundary frames with siamese contexts and supplements them with the neighboring query-related visual information. Therefore, it brings significant improvement by using the soft label in training process.

\begin{table}[t!]
    \small
    \centering
    \setlength{\tabcolsep}{1.8mm}{
    \begin{tabular}{c|cccc}
    \toprule 
    Number & K=1 & K=2 & K=4 & K=8 \\
    \midrule
    R@1, IoU=0.5 & 50.45 & 52.10 & 54.49 & \textbf{54.62} \\
    R@1, IoU=0.7 & 29.64 & 30.78 & 33.15 & \textbf{33.27} \\
    \bottomrule
    \end{tabular}}
    \caption{The effect of the number $K$ of the sampled siamese frames on ActivityNet Captions dataset.}
    \label{tab:ablation2}
    \vspace{-10pt}
\end{table}

\noindent \textbf{Effect of the number of siamese frames}~~
We compare our method with various number of siamese frames as shown in Table~\ref{tab:ablation2}. 
When adding the siamese sample number $K$ from 1 to 8, our method dynamically promotes the accuracy. Such improvement can demonstrate that more siamese samples can bring richer knowledge, which makes our network benefited from it. Although the accuracy is increasing with the number of siamese frames, we observe that the improvement from the number 4 to 8 is slight. We think the reason is the saturation of knowledge, \textit{i.e.}, the model has enough knowledge to learn the task on this dataset. Hence, it is almost meaningless to purely increase the siamese frames. To balance the training time and accuracy, we assign $K$ = 4 in our final version.

\noindent \textbf{Plug-and-Play}~~
Our proposed siamese learning strategy is flexible and can be adopted to other TSG methods for anchor feature enhancement. As shown in Table \ref{tab:ablation3}, we directly apply siamese learning strategy into existing module for anchor feature enriching \textit{without} using soft label training. It shows that our siamese learning strategy can provide more contextual and fine-grained information for anchor feature encoding, bringing large improvement.

\begin{table}[t!]
    \small
    \centering
    \setlength{\tabcolsep}{1.2mm}{
    \begin{tabular}{c|c|cc}
    \toprule
    \multirow{2}*{Methods} & \multirow{2}*{Variant} & R@1, & R@1, \\ 
    ~ & ~ & IoU=0.5 & IoU=0.7 \\  \midrule
    \multirow{2}*{VSLNet} & Origin & 43.22 & 26.16  \\
    ~ & +siamese & \textbf{50.38} & \textbf{30.06} \\ \midrule
    \multirow{2}*{CBLN} & Origin & 48.12 & 27.60  \\
    ~ & +siamese & \textbf{56.86} & \textbf{30.79}  \\ \midrule
    \multirow{2}*{MGSL} & Origin & 51.87 & 31.42  \\
    ~ & +siamese & \textbf{58.77} & \textbf{33.41} \\ \bottomrule
    \end{tabular}}
    \caption{We apply our siamese learning strategy to existing TSG models on ActivityNet Captions dataset.}
    \label{tab:ablation3}
    \vspace{-10pt}
\end{table}

\subsection{Qualitative Results}
In Figure~\ref{fig:result}, we show two visualization examples to qualitatively analyze what kind of knowledge does the siamese frames bring to the anchor frames. It is unavoidable to lose some visual contents when sparsely sampling from the video. Especially for the boundary frames that are easily to be filtered out by sampling, the visual content of the newly sampled boundary may lose query-relevant information (\textit{e.g.}, \textcolor{brown}{brown} words in figure). However, we can obtain the absent contents from their siamese frames due to different sampling indices and duration. Hence, our siamese frames can enrich and supplement the sampled frames with more consecutive query-related visual semantics to make a fine-grained video comprehension, keeping the appropriate segment length of the sampled video for more accurate boundary prediction. 

\section{Conclusion}
In this paper, we propose a novel Siamese Sampling and Reasoning Network (SSRN) to alleviate the limitations of both boundary-bias and reasoning-bias in existing TSG methods. In addition to the original anchor frames, our model also samples a certain number of siamese frames from the same video to enrich and refine the visual semantics of the anchor frames. A soft label is further exploited to supervise the enhanced anchor features for predicting more accurate segment boundaries. Experimental results show both effectiveness and efficiency of our SSRN on three challenging datasets.

\section*{Limitations}
This work analyzes an interesting problem of how to learn from inside to address the limitation of the boundary-bias on the temporal sentence grounding. Since our method targets on the issue of long video sampling, it may be not helpful to handle the short video processing but still can improve the contextual representation learning for the short video. Besides, our sampled siamese frames would bring extra burden (\textit{e.g.}, computation, memory and parameters) during the training and testing. Therefore, a more light way to ease the siamese knowledge extraction is a promising future direction.

\section{Acknowledgments}
This work was supported by National Natural Science Foundation of China (No.61972448, No.62272328, No.62172038 and No.62172068).

\bibliography{anthology,custom}

\begin{thebibliography}{51}
\expandafter\ifx\csname natexlab\endcsname\relax\def\natexlab#1{#1}\fi

\bibitem[{Anne~Hendricks et~al.(2017)Anne~Hendricks, Wang, Shechtman, Sivic,
  Darrell, and Russell}]{anne2017localizing}
Lisa Anne~Hendricks, Oliver Wang, Eli Shechtman, Josef Sivic, Trevor Darrell,
  and Bryan Russell. 2017.
\newblock Localizing moments in video with natural language.
\newblock In \emph{Proceedings of the IEEE International Conference on Computer
  Vision (ICCV)}.

\bibitem[{Cao et~al.(2021)Cao, Chen, Shou, Zhang, and Zou}]{cao2021pursuit}
Meng Cao, Long Chen, Mike~Zheng Shou, Can Zhang, and Yuexian Zou. 2021.
\newblock On pursuit of designing multi-modal transformer for video grounding.
\newblock In \emph{Proceedings of the 2021 Conference on Empirical Methods in
  Natural Language Processing (EMNLP)}, pages 9810--9823.

\bibitem[{Carreira and Zisserman(2017)}]{carreira2017quo}
Joao Carreira and Andrew Zisserman. 2017.
\newblock Quo vadis, action recognition? a new model and the kinetics dataset.
\newblock In \emph{proceedings of the IEEE Conference on Computer Vision and
  Pattern Recognition (CVPR)}, pages 6299--6308.

\bibitem[{Chen et~al.(2018)Chen, Chen, Ma, Jie, and Chua}]{chen2018temporally}
Jingyuan Chen, Xinpeng Chen, Lin Ma, Zequn Jie, and Tat-Seng Chua. 2018.
\newblock Temporally grounding natural sentence in video.
\newblock In \emph{Proceedings of the 2018 Conference on Empirical Methods in
  Natural Language Processing (EMNLP)}, pages 162--171.

\bibitem[{Chen et~al.(2020)Chen, Lu, Tang, Xiao, Zhang, Tan, and
  Li}]{chenrethinking}
Long Chen, Chujie Lu, Siliang Tang, Jun Xiao, Dong Zhang, Chilie Tan, and
  Xiaolin Li. 2020.
\newblock Rethinking the bottom-up framework for query-based video
  localization.
\newblock In \emph{Proceedings of the AAAI Conference on Artificial
  Intelligence}.

\bibitem[{Chung et~al.(2014)Chung, Gulcehre, Cho, and
  Bengio}]{chung2014empirical}
Junyoung Chung, Caglar Gulcehre, KyungHyun Cho, and Yoshua Bengio. 2014.
\newblock Empirical evaluation of gated recurrent neural networks on sequence
  modeling.
\newblock In \emph{Advances in Neural Information Processing Systems (NIPS)}.

\bibitem[{Dong et~al.(2019)Dong, Li, Xu, Ji, and Wang}]{2019Dual}
Jianfeng Dong, Xirong Li, Chaoxi Xu, Shouling Ji, and Xun Wang. 2019.
\newblock Dual encoding for zero-example video retrieval.
\newblock In \emph{Proceedings of the IEEE Conference on Computer Vision and
  Pattern Recognition (CVPR)}.

\bibitem[{Fang et~al.(2022)Fang, Liu, Zhou, and Hu}]{fang2022multi}
Xiang Fang, Daizong Liu, Pan Zhou, and Yuchong Hu. 2022.
\newblock Multi-modal cross-domain alignment network for video moment
  retrieval.
\newblock \emph{IEEE Transactions on Multimedia}.

\bibitem[{Gao et~al.(2017)Gao, Sun, Yang, and Nevatia}]{gao2017tall}
Jiyang Gao, Chen Sun, Zhenheng Yang, and Ram Nevatia. 2017.
\newblock Tall: Temporal activity localization via language query.
\newblock In \emph{Proceedings of the IEEE International Conference on Computer
  Vision (ICCV)}, pages 5267--5275.

\bibitem[{Ge et~al.(2019)Ge, Gao, Chen, and Nevatia}]{ge2019mac}
Runzhou Ge, Jiyang Gao, Kan Chen, and Ram Nevatia. 2019.
\newblock Mac: Mining activity concepts for language-based temporal
  localization.
\newblock In \emph{IEEE Winter Conference on Applications of Computer Vision
  (WACV)}, pages 245--253.

\bibitem[{Krishna et~al.(2017)Krishna, Hata, Ren, Fei-Fei, and
  Carlos~Niebles}]{krishna2017dense}
Ranjay Krishna, Kenji Hata, Frederic Ren, Li~Fei-Fei, and Juan Carlos~Niebles.
  2017.
\newblock Dense-captioning events in videos.
\newblock In \emph{Proceedings of the IEEE International Conference on Computer
  Vision (ICCV)}, pages 706--715.

\bibitem[{Liu et~al.(2022{\natexlab{a}})Liu, Fang, Hu, and
  Zhou}]{liu2022exploringa}
Daizong Liu, Xiang Fang, Wei Hu, and Pan Zhou. 2022{\natexlab{a}}.
\newblock Exploring optical-flow-guided motion and detection-based appearance
  for temporal sentence grounding.
\newblock \emph{arXiv preprint arXiv:2203.02966}.

\bibitem[{Liu and Hu(2022{\natexlab{a}})}]{liu2022learning}
Daizong Liu and Wei Hu. 2022{\natexlab{a}}.
\newblock Learning to focus on the foreground for temporal sentence grounding.
\newblock In \emph{Proceedings of the 29th International Conference on
  Computational Linguistics}, pages 5532--5541.

\bibitem[{Liu and Hu(2022{\natexlab{b}})}]{liu2022skimming}
Daizong Liu and Wei Hu. 2022{\natexlab{b}}.
\newblock Skimming, locating, then perusing: A human-like framework for natural
  language video localization.
\newblock In \emph{Proceedings of the 30th ACM International Conference on
  Multimedia}, pages 4536--4545.

\bibitem[{Liu et~al.(2022{\natexlab{b}})Liu, Qu, Di, Cheng, Xu, and
  Zhou}]{liu2022memory}
Daizong Liu, Xiaoye Qu, Xing Di, Yu~Cheng, Zichuan~Xu Xu, and Pan Zhou.
  2022{\natexlab{b}}.
\newblock Memory-guided semantic learning network for temporal sentence
  grounding.
\newblock In \emph{Proceedings of the AAAI Conference on Artificial
  Intelligence}.

\bibitem[{Liu et~al.(2020{\natexlab{a}})Liu, Qu, Dong, and
  Zhou}]{liu2020reasoning}
Daizong Liu, Xiaoye Qu, Jianfeng Dong, and Pan Zhou. 2020{\natexlab{a}}.
\newblock Reasoning step-by-step: Temporal sentence localization in videos via
  deep rectification-modulation network.
\newblock In \emph{Proceedings of the 28th International Conference on
  Computational Linguistics}, pages 1841--1851.

\bibitem[{Liu et~al.(2021{\natexlab{a}})Liu, Qu, Dong, and
  Zhou}]{liu2021adaptive}
Daizong Liu, Xiaoye Qu, Jianfeng Dong, and Pan Zhou. 2021{\natexlab{a}}.
\newblock Adaptive proposal generation network for temporal sentence
  localization in videos.
\newblock In \emph{Proceedings of the 2021 Conference on Empirical Methods in
  Natural Language Processing (EMNLP)}, pages 9292--9301.

\bibitem[{Liu et~al.(2021{\natexlab{b}})Liu, Qu, Dong, Zhou, Cheng, Wei, Xu,
  and Xie}]{liu2021context}
Daizong Liu, Xiaoye Qu, Jianfeng Dong, Pan Zhou, Yu~Cheng, Wei Wei, Zichuan Xu,
  and Yulai Xie. 2021{\natexlab{b}}.
\newblock Context-aware biaffine localizing network for temporal sentence
  grounding.
\newblock In \emph{Proceedings of the IEEE/CVF Conference on Computer Vision
  and Pattern Recognition}.

\bibitem[{Liu et~al.(2022{\natexlab{c}})Liu, Qu, and Hu}]{liu2022reducing}
Daizong Liu, Xiaoye Qu, and Wei Hu. 2022{\natexlab{c}}.
\newblock Reducing the vision and language bias for temporal sentence
  grounding.
\newblock In \emph{Proceedings of the 30th ACM International Conference on
  Multimedia}, pages 4092--4101.

\bibitem[{Liu et~al.(2020{\natexlab{b}})Liu, Qu, Liu, Dong, Zhou, and
  Xu}]{liu2020jointly}
Daizong Liu, Xiaoye Qu, Xiao-Yang Liu, Jianfeng Dong, Pan Zhou, and Zichuan Xu.
  2020{\natexlab{b}}.
\newblock Jointly cross-and self-modal graph attention network for query-based
  moment localization.
\newblock In \emph{Proceedings of the 28th ACM International Conference on
  Multimedia}, pages 4070--4078.

\bibitem[{Liu et~al.(2022{\natexlab{d}})Liu, Qu, Wang, Di, Zou, Cheng, Xu, and
  Zhou}]{liu2022unsupervised}
Daizong Liu, Xiaoye Qu, Yinzhen Wang, Xing Di, Kai Zou, Yu~Cheng, Zichuan Xu,
  and Pan Zhou. 2022{\natexlab{d}}.
\newblock Unsupervised temporal video grounding with deep semantic clustering.
\newblock In \emph{Proceedings of the AAAI Conference on Artificial
  Intelligence}.

\bibitem[{Liu et~al.(2021{\natexlab{c}})Liu, Qu, and
  Zhou}]{liu2021progressively}
Daizong Liu, Xiaoye Qu, and Pan Zhou. 2021{\natexlab{c}}.
\newblock Progressively guide to attend: An iterative alignment framework for
  temporal sentence grounding.
\newblock In \emph{Proceedings of the 2021 Conference on Empirical Methods in
  Natural Language Processing (EMNLP)}, pages 9302--9311.

\bibitem[{Liu et~al.(2022{\natexlab{e}})Liu, Qu, Zhou, and
  Liu}]{liu2022exploring}
Daizong Liu, Xiaoye Qu, Pan Zhou, and Yang Liu. 2022{\natexlab{e}}.
\newblock Exploring motion and appearance information for temporal sentence
  grounding.
\newblock In \emph{Proceedings of the AAAI Conference on Artificial
  Intelligence}.

\bibitem[{Liu et~al.(2022{\natexlab{f}})Liu, Zhou, Xu, Wang, and
  Li}]{liu2022few}
Daizong Liu, Pan Zhou, Zichuan Xu, Haozhao Wang, and Ruixuan Li.
  2022{\natexlab{f}}.
\newblock Few-shot temporal sentence grounding via memory-guided semantic
  learning.
\newblock \emph{IEEE Transactions on Circuits and Systems for Video
  Technology}.

\bibitem[{Liu et~al.(2018{\natexlab{a}})Liu, Wang, Nie, He, Chen, and
  Chua}]{liu2018attentive}
Meng Liu, Xiang Wang, Liqiang Nie, Xiangnan He, Baoquan Chen, and Tat-Seng
  Chua. 2018{\natexlab{a}}.
\newblock Attentive moment retrieval in videos.
\newblock In \emph{Proceedings of the 41nd International ACM SIGIR Conference
  on Research and Development in Information Retrieval (SIGIR)}, pages 15--24.

\bibitem[{Liu et~al.(2018{\natexlab{b}})Liu, Wang, Nie, Tian, Chen, and
  Chua}]{liu2018cross}
Meng Liu, Xiang Wang, Liqiang Nie, Qi~Tian, Baoquan Chen, and Tat-Seng Chua.
  2018{\natexlab{b}}.
\newblock Cross-modal moment localization in videos.
\newblock In \emph{Proceedings of the 26th ACM international conference on
  Multimedia}, pages 843--851.

\bibitem[{Lu et~al.(2019)Lu, Chen, Tan, Li, and Xiao}]{lu-etal-2019-debug}
Chujie Lu, Long Chen, Chilie Tan, Xiaolin Li, and Jun Xiao. 2019.
\newblock {DEBUG}: A dense bottom-up grounding approach for natural language
  video localization.
\newblock In \emph{Proceedings of the 2019 Conference on Empirical Methods in
  Natural Language Processing and the 9th International Joint Conference on
  Natural Language Processing (EMNLP-IJCNLP)}.

\bibitem[{Mun et~al.(2020)Mun, Cho, and Han}]{mun2020local}
Jonghwan Mun, Minsu Cho, and Bohyung Han. 2020.
\newblock Local-global video-text interactions for temporal grounding.
\newblock In \emph{Proceedings of the IEEE Conference on Computer Vision and
  Pattern Recognition (CVPR)}, pages 10810--10819.

\bibitem[{Nan et~al.(2021)Nan, Qiao, Xiao, Liu, Leng, Zhang, and
  Lu}]{nan2021interventional}
Guoshun Nan, Rui Qiao, Yao Xiao, Jun Liu, Sicong Leng, Hao Zhang, and Wei Lu.
  2021.
\newblock Interventional video grounding with dual contrastive learning.
\newblock In \emph{Proceedings of the IEEE Conference on Computer Vision and
  Pattern Recognition (CVPR)}.

\bibitem[{Pennington et~al.(2014)Pennington, Socher, and
  Manning}]{pennington2014glove}
Jeffrey Pennington, Richard Socher, and Christopher~D Manning. 2014.
\newblock Glove: Global vectors for word representation.
\newblock In \emph{Proceedings of the Conference on Empirical Methods in
  Natural Language Processing (EMNLP)}, pages 1532--1543.

\bibitem[{Qu et~al.(2020)Qu, Tang, Zou, Cheng, Dong, Zhou, and Xu}]{qu2020fine}
Xiaoye Qu, Pengwei Tang, Zhikang Zou, Yu~Cheng, Jianfeng Dong, Pan Zhou, and
  Zichuan Xu. 2020.
\newblock Fine-grained iterative attention network for temporal language
  localization in videos.
\newblock In \emph{Proceedings of the 28th ACM International Conference on
  Multimedia}, pages 4280--4288.

\bibitem[{Regneri et~al.(2013)Regneri, Rohrbach, Wetzel, Thater, Schiele, and
  Pinkal}]{regneri2013grounding}
Michaela Regneri, Marcus Rohrbach, Dominikus Wetzel, Stefan Thater, Bernt
  Schiele, and Manfred Pinkal. 2013.
\newblock Grounding action descriptions in videos.
\newblock \emph{Transactions of the Association for Computational Linguistics},
  1:25--36.

\bibitem[{Rodriguez et~al.(2020)Rodriguez, Marrese-Taylor, Saleh, Li, and
  Gould}]{rodriguez2020proposal}
Cristian Rodriguez, Edison Marrese-Taylor, Fatemeh~Sadat Saleh, Hongdong Li,
  and Stephen Gould. 2020.
\newblock Proposal-free temporal moment localization of a natural-language
  query in video using guided attention.
\newblock In \emph{The IEEE Winter Conference on Applications of Computer
  Vision (WACV)}, pages 2464--2473.

\bibitem[{Sigurdsson et~al.(2016)Sigurdsson, Varol, Wang, Farhadi, Laptev, and
  Gupta}]{sigurdsson2016hollywood}
Gunnar~A Sigurdsson, G{\"u}l Varol, Xiaolong Wang, Ali Farhadi, Ivan Laptev,
  and Abhinav Gupta. 2016.
\newblock Hollywood in homes: Crowdsourcing data collection for activity
  understanding.
\newblock In \emph{European Conference on Computer Vision (ECCV)}, pages
  510--526.

\bibitem[{Simonyan and Zisserman(2014)}]{simonyan2014very}
Karen Simonyan and Andrew Zisserman. 2014.
\newblock Very deep convolutional networks for large-scale image recognition.
\newblock \emph{arXiv preprint arXiv:1409.1556}.

\bibitem[{Singha et~al.(2018)Singha, Roy, and Laskar}]{singha2018dynamic}
Joyeeta Singha, Amarjit Roy, and Rabul~Hussain Laskar. 2018.
\newblock Dynamic hand gesture recognition using vision-based approach for
  human--computer interaction.
\newblock \emph{Neural Computing and Applications}.

\bibitem[{Tran et~al.(2015)Tran, Bourdev, Fergus, Torresani, and
  Paluri}]{tran2015learning}
Du~Tran, Lubomir Bourdev, Rob Fergus, Lorenzo Torresani, and Manohar Paluri.
  2015.
\newblock Learning spatiotemporal features with 3d convolutional networks.
\newblock In \emph{Proceedings of the IEEE International Conference on Computer
  Vision (ICCV)}, pages 4489--4497.

\bibitem[{Vaswani et~al.(2017)Vaswani, Shazeer, Parmar, Uszkoreit, Jones,
  Gomez, Kaiser, and Polosukhin}]{vaswani2017attention}
Ashish Vaswani, Noam Shazeer, Niki Parmar, Jakob Uszkoreit, Llion Jones,
  Aidan~N Gomez, {\L}ukasz Kaiser, and Illia Polosukhin. 2017.
\newblock Attention is all you need.
\newblock In \emph{Advances in Neural Information Processing Systems (NIPS)},
  pages 5998--6008.

\bibitem[{Wang et~al.(2020)Wang, Ma, and Jiang}]{wang2019temporally}
Jingwen Wang, Lin Ma, and Wenhao Jiang. 2020.
\newblock Temporally grounding language queries in videos by contextual
  boundary-aware prediction.
\newblock In \emph{Proceedings of the AAAI Conference on Artificial
  Intelligence}.

\bibitem[{Wang et~al.(2019)Wang, Huang, and Wang}]{wang2019language}
Weining Wang, Yan Huang, and Liang Wang. 2019.
\newblock Language-driven temporal activity localization: A semantic matching
  reinforcement learning model.
\newblock In \emph{Proceedings of the IEEE Conference on Computer Vision and
  Pattern Recognition (CVPR)}, pages 334--343.

\bibitem[{Xiao et~al.(2021)Xiao, Chen, Shao, Zhuang, and
  Xiao}]{xiao2021natural}
Shaoning Xiao, Long Chen, Jian Shao, Yueting Zhuang, and Jun Xiao. 2021.
\newblock Natural language video localization with learnable moment proposals.
\newblock In \emph{Proceedings of the 2021 Conference on Empirical Methods in
  Natural Language Processing}, pages 4008--4017.

\bibitem[{Xu et~al.(2019)Xu, He, Plummer, Sigal, Sclaroff, and
  Saenko}]{xu2019multilevel}
Huijuan Xu, Kun He, Bryan~A Plummer, Leonid Sigal, Stan Sclaroff, and Kate
  Saenko. 2019.
\newblock Multilevel language and vision integration for text-to-clip
  retrieval.
\newblock In \emph{Proceedings of the AAAI Conference on Artificial
  Intelligence}, volume~33, pages 9062--9069.

\bibitem[{Yang et~al.(2020)Yang, Dong, Cao, Wang, Wang, and
  Chua}]{yang2020tree}
Xun Yang, Jianfeng Dong, Yixin Cao, Xun Wang, Meng Wang, and Tat-Seng Chua.
  2020.
\newblock Tree-augmented cross-modal encoding for complex-query video
  retrieval.
\newblock In \emph{Proceedings of the 43rd International ACM SIGIR Conference
  on Research and Development in Information Retrieval (SIGIR)}, pages
  1339--1348.

\bibitem[{Yuan et~al.(2019{\natexlab{a}})Yuan, Ma, Wang, Liu, and
  Zhu}]{yuan2019semantic}
Yitian Yuan, Lin Ma, Jingwen Wang, Wei Liu, and Wenwu Zhu. 2019{\natexlab{a}}.
\newblock Semantic conditioned dynamic modulation for temporal sentence
  grounding in videos.
\newblock In \emph{Advances in Neural Information Processing Systems (NIPS)},
  pages 534--544.

\bibitem[{Yuan et~al.(2019{\natexlab{b}})Yuan, Mei, and Zhu}]{yuan2019find}
Yitian Yuan, Tao Mei, and Wenwu Zhu. 2019{\natexlab{b}}.
\newblock To find where you talk: Temporal sentence localization in video with
  attention based location regression.
\newblock In \emph{Proceedings of the AAAI Conference on Artificial
  Intelligence}, volume~33, pages 9159--9166.

\bibitem[{Zeng et~al.(2020)Zeng, Xu, Huang, Chen, Tan, and Gan}]{zeng2020dense}
Runhao Zeng, Haoming Xu, Wenbing Huang, Peihao Chen, Mingkui Tan, and Chuang
  Gan. 2020.
\newblock Dense regression network for video grounding.
\newblock In \emph{Proceedings of the IEEE Conference on Computer Vision and
  Pattern Recognition (CVPR)}, pages 10287--10296.

\bibitem[{Zhang et~al.(2019{\natexlab{a}})Zhang, Dai, Wang, Wang, and
  Davis}]{zhang2019man}
Da~Zhang, Xiyang Dai, Xin Wang, Yuan-Fang Wang, and Larry~S Davis.
  2019{\natexlab{a}}.
\newblock Man: Moment alignment network for natural language moment retrieval
  via iterative graph adjustment.
\newblock In \emph{Proceedings of the IEEE Conference on Computer Vision and
  Pattern Recognition (CVPR)}, pages 1247--1257.

\bibitem[{Zhang et~al.(2021)Zhang, Sun, Jing, Zhen, and Goh}]{2021Parallel}
H.~Zhang, A.~Sun, W.~Jing, L.~Zhen, and Rsm Goh. 2021.
\newblock Parallel attention network with sequence matching for video
  grounding.
\newblock In \emph{Findings of the Association for Computational Linguistics:
  ACL-IJCNLP 2021}.

\bibitem[{Zhang et~al.(2020{\natexlab{a}})Zhang, Sun, Jing, and
  Zhou}]{zhang2020span}
Hao Zhang, Aixin Sun, Wei Jing, and Joey~Tianyi Zhou. 2020{\natexlab{a}}.
\newblock Span-based localizing network for natural language video
  localization.
\newblock In \emph{Proceedings of the 58th Annual Meeting of the Association
  for Computational Linguistics}, pages 6543--6554.

\bibitem[{Zhang et~al.(2020{\natexlab{b}})Zhang, Peng, Fu, and
  Luo}]{zhang2019learning}
Songyang Zhang, Houwen Peng, Jianlong Fu, and Jiebo Luo. 2020{\natexlab{b}}.
\newblock Learning 2d temporal adjacent networks for moment localization with
  natural language.
\newblock In \emph{Proceedings of the AAAI Conference on Artificial
  Intelligence}.

\bibitem[{Zhang et~al.(2019{\natexlab{b}})Zhang, Lin, Zhao, and
  Xiao}]{zhang2019cross}
Zhu Zhang, Zhijie Lin, Zhou Zhao, and Zhenxin Xiao. 2019{\natexlab{b}}.
\newblock Cross-modal interaction networks for query-based moment retrieval in
  videos.
\newblock In \emph{Proceedings of the 42nd International ACM SIGIR Conference
  on Research and Development in Information Retrieval (SIGIR)}, pages
  655--664.

\end{thebibliography}
\bibliographystyle{acl_natbib}




\end{document}